\documentclass[acmsmall,screen,nonacm]{acmart}

\usepackage{epsfig}
\usepackage{graphicx}
\usepackage{amsmath}
\usepackage{epstopdf}
\usepackage{color}
\usepackage{array}
\usepackage{comment}
\usepackage{subcaption}
\makeatletter  
\newif\if@restonecol  
\makeatother

\usepackage{multirow}
\usepackage{algorithm}
\usepackage{algorithmic}

\usepackage{newtxmath}

\AtBeginDocument{%
    \providecommand\BibTeX{{%
            \normalfont B\kern-0.5em{\scshape i\kern-0.25em b}\kern-0.8em\TeX}}}

\setcopyright{none}
\graphicspath{{./figure/}}
\newcommand{\etal}{\textit{et al.}}
\newcommand{\eg}{\textit{e.g.}}

\newcommand{\etc}{\textit{etc.}}

\begin{document}
    
\title{Geometry Regularization for Vanilla NeRF via Self-Supervised Cross-Ray Consistency}

\author{Zongji Wang}
\email{wangzongji@aircas.ac.cn}
\authorsaddresses{}

\renewcommand{\shortauthors}{Wang}

\begin{abstract}
    Neural Radiance Fields (NeRF) achieve high-quality novel-view synthesis by representing a scene as a continuous volumetric radiance field. However, the geometry encoded by the density field of vanilla NeRF can be noisy and inaccurate, since different combinations of shape and radiance may explain the same input images. To reduce this ambiguity, we augment vanilla NeRF with a geometry-optimization branch based on self-supervised cross-ray consistency. Specifically, this branch samples candidate surface points from the density field, generates virtual rays that intersect at each point from different directions, and renders their appearance using the NeRF network. 
    Based on the visibility-weighted color consistency across these rays, we constrain the implicit geometry.
    We further regularize the implicit geometry by penalizing opacity at candidates with inconsistent appearance predictions and encouraging local normal smoothness. These geometric constraints are derived from the NeRF predictions and require no ground-truth depth or surface-normal supervision.
    Experiments on eight Replica scenes show that the proposed method reduces the mean Chamfer distance by 30.8\% compared with vanilla NeRF.
\end{abstract}


\maketitle

\section{Introduction} \label{sec:Introduction}
Recovering 3D shape from 2D observations is a cornerstone problem of computer vision and graphics. This long-standing research topic supports various applications in AR, VR, games, movies, \etc~\cite{NeRFShape:Azinovic2022CVPR, NeRFShape:boss2022samurai}
With the advances of inverse graphics,
differentiable renderer can be integrated in neural network pipelines, making it feasible to directly optimize the 3D geometric parameters using observed 2D images.
Among various 3D geometry representations, implicit function represents 3D content continuously and without discretization. With its small memory footprint, implicit neural representation is popular in recent studies. One of the most successful approaches is NeRF (Neural Radiance Fields)~\cite{NeRF:Mildenhall2021}.

NeRF is a scene representation method presented by Mildenhall \etal~\cite{NeRF:Mildenhall2021}. In that work, a deep neural network (typically, a multilayer perceptron, MLP) is applied to parameterize a continuous implicit function, which takes the position $x \in \mathbb{R}^3$ and viewing direction $d \in \mathbb{R}^3$ as input, and predicts the density $\sigma(x) \in \mathbb{R}^+$ and view-dependent color $c(x,d) \in \mathbb{R}^3$. Through differentiable volume rendering, dense colors from multiple 2D images can be used directly to optimize the neural radiance field. Encoding the information from multi-view 2D images, this powerful representation shows promising results in capturing complex scene structures and synthesizing 3D-consistent images with fine details.
The superior rendering effects ignite research enthusiasm towards NeRF-like methods.
Subsequent works include modification for
dynamic scene modelling~\cite{NeRFDynamic:li2021neural, NeRFDynamic:fang2022fast}, 
large-scale scene representation~\cite{NeRFLarge:tancik2022block,NeRFLarge:turki2022mega,NeRFLarge:xiangli2022bungee}, 
baking approaches for real-time rendering~\cite{NeRFBake:hedman2021snerg,NeRFBake:chen2022mobile,NeRFBake:yariv2023bakedsdf},  
encoding extra layers for scene understanding (\eg~ intrinsic~\cite{NeRFLayer:ye2022Intrinsic} and semantic~\cite{NeRFLayer:zhi2021inplace} layers), 
and 3D-aware content or scene generation~\cite{NeRFGAN:gu2021style, NeRFGAN:deng2022gram, NeRFGAN:chen2023sd}. 
Although these approaches have shown impressive results for novel view synthesis, the 3D geometry extracted from the underlying volume density is far from satisfactory, usually suffering from noise and inaccuracy~\cite{NeRFShape:yariv2021volume, NeRFBake:yariv2023bakedsdf}.

\begin{figure}[htbp]
    \begin{center}
        \includegraphics[width=\linewidth]{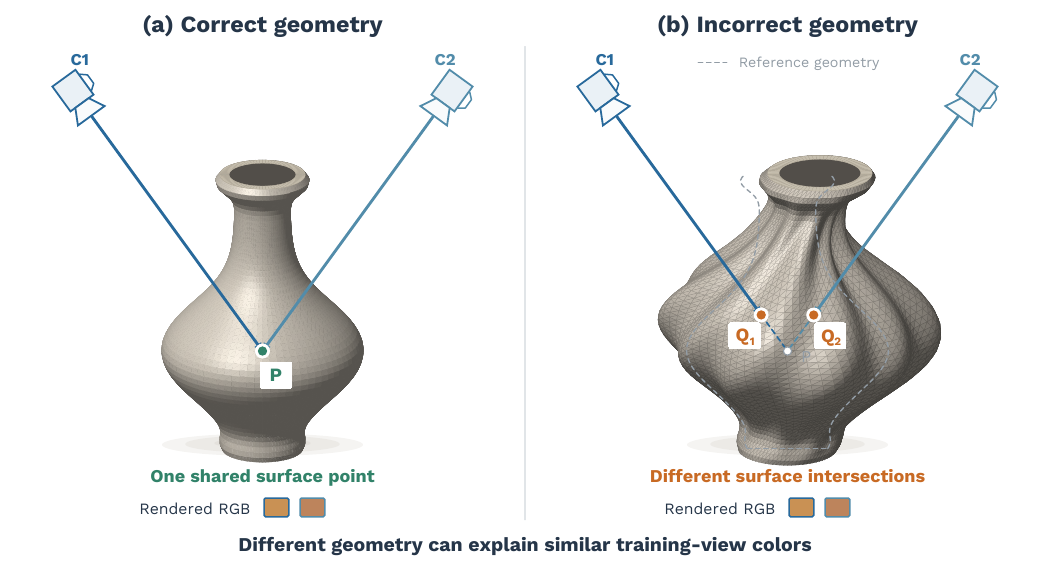}
    \end{center}
    \caption{Schematic illustration of geometry--appearance ambiguity. (a) Two captured rays observe the same physical surface point $P$ on a coherent 3D object. (b) An inflated, distorted surface intercepts the same rays at different points $Q_1$ and $Q_2$. Based on view-dependent radiance, the incorrect geometry can nevertheless reproduce similar rendered RGB values along each captured ray, as indicated by the paired color swatches. Thus, image agreement alone does not uniquely determine the geometry. The dashed contour denotes the reference geometry, and the dashed ray segments extend toward $P$ behind the incorrect surface.}
    \Description{An original three-dimensional schematic showing two vase-shaped surface hypotheses under the same two camera viewpoints. The correct surface is smooth, and both rays meet at a shared point P. The incorrect surface is inflated and distorted, with a visible triangular mesh; the rays intersect it at distinct points Q1 and Q2. Paired color swatches indicate similar rendered RGB values for corresponding rays despite the geometry difference. A dashed reference contour and dashed ray extensions show the underlying reference shape and point.}
    \label{fig:idea}
\end{figure}

Neural implicit functions help circumvent the limitations about complex pipelines in classic depth-based multi-view stereo (MVS)~\cite{MVS:schonberger2016pixelwise} and expensive memory consumption in voxel-based approaches~\cite{Voxel:ulusoy2015towards,Voxel:wang2019voxsegnet}.
However, it is still a fundamentally under-constrained problem to recover accurate geometry from 2D images, even for approaches based on differentiable rendering~\cite{NeRF:barron2022mip}. One of the main observations is the ambiguity between 3D shape and radiance, dubbed shape-radiance ambiguity~\cite{NeRF:zhang2020nerf++}. As illustrated in Figure~\ref{fig:idea}, different surface hypotheses can reproduce similar training-view colors through view-dependent radiance. This ambiguity makes it difficult to recover the scene's actual geometry from image agreement alone~\cite{NeRF:zhang2020nerf++, NeRF:barron2022mip, Carving:kutulakos2000theory}.

A lot of effort has been dedicated to improve the 3D reconstruction quality of implicit neural representations.
To put it briefly, 
there are works applying signed distance function (SDF) to model the 3D surface~\cite{NeRFShape:niemeyer2020DVR, NeRFShape:yariv2021volume, NeRFShape:yariv2020multiview, NeRFShape:wang2021neus, NeRFBake:yariv2023bakedsdf}, 
combining neural implicit field with explicit 3D representations such as voxel~\cite{NeRFVox:liu2020neural, NeRFVox:yang2021object} and polygon mesh~\cite{NeRFBake:chen2022mobile, NeRFTet:munkberg2022extracting},
optimizing sampling approach of ray marching~\cite{NeRFVox:liu2020neural, NeRFShape:Oechsle2021unisurf, NeRFShape:wang2021neus, NeRFShape:rematas2022urban},
applying regularizers about smoothness and sparsity to the implicit field~\cite{NeRF:Mildenhall2021, NeRFShape:Oechsle2021unisurf, NeRFRender:yu2021plenoctrees},
adding priors about depth and structure to constrain the geometry~\cite{NeRFShape:Oechsle2021unisurf, NeRFMVS:wei2021nerfingmvs, NeRFShape:yu2022monosdf, NeRFShape:guo2022manhattan},
and using multi-view consistency as MVS-based reconstruction~\cite{NeRFMVS:darmon2022improving, NeRFMVS:wei2021nerfingmvs, NeRFMVS:chen2021mvsnerf}.

In this paper, we directly refine the implicit geometry of vanilla NeRF to reduce its shape-radiance ambiguity. We choose vanilla NeRF~\cite{NeRF:Mildenhall2021} because its simple radiance-field representation supports complex scenes and arbitrary camera poses. However, as its geometry is represented only by volume density, the extracted surfaces often contain noise and inaccuracies.
To improve the geometric fidelity, we build on the observation that the appearance of a surface point cannot vary arbitrarily across nearby viewpoints~\cite{Carving:kutulakos2000theory}. Existing neural reconstruction methods use this photo-consistency cue through feature matching or image warping between captured views~\cite{NeRFMVS:darmon2022improving, NeRFMVS:wei2021nerfingmvs, NeRFMVS:chen2021mvsnerf}. We instead extend the NeRF architecture with a branch that samples candidate surface points, generates virtual viewpoints around them, and renders the corresponding cross-ray appearances. Based on these predictions, we obtain a self-supervised signal for optimizing the density field.

\textbf{Contributions.} We summarize the contributions as follows:

\begin{itemize}
    \item We design a geometry-optimization branch for vanilla NeRF that derives geometric cues from the density field and integrates multi-view appearance constraints into the differentiable rendering pipeline.

    \item We propose visibility-weighted, self-supervised cross-ray consistency and penalize opacity at candidates whose rendered appearances disagree across virtual views.

    \item Experiments on eight Replica scenes show a 30.8\% reduction in mean Chamfer distance relative to vanilla NeRF, demonstrating the effectiveness of the proposed framework in improving implicit geometry.

\end{itemize}

\section{Related Work}

\subsection{3D Geometric Representations}

3D geometry is commonly represented by volumes, point clouds, meshes, or implicit functions. Volumetric representations encode spatial structure directly~\cite{Voxel:wang2019voxsegnet}, but their memory consumption often limits resolution. Point clouds are compact and flexible, although they lack an explicit neighborhood structure. Meshes are compatible with conventional graphics pipelines~\cite{NeRFBake:yariv2023bakedsdf}, but changes in topology require modifying their connectivity. Implicit representations instead describe geometry through continuous functions, such as occupancy fields or signed distance fields (SDFs)~\cite{NeRFShape:niemeyer2020DVR,NeRFShape:yariv2021volume}. Neural networks parameterize these functions without maintaining a dense spatial grid. We focus on the geometry encoded by implicit neural representations.

\subsection{Implicit Neural Representations}

NeRF~\cite{NeRF:Mildenhall2021} represents a scene with a neural network that maps 3D positions and viewing directions to density and view-dependent color. Through differentiable volume rendering, it learns the radiance field directly from posed images. Its continuous representation supports novel-view synthesis without explicitly reconstructing a surface during training.

Subsequent work extends this representation to dynamic scenes~\cite{NeRFDynamic:li2021neural,NeRFDynamic:fang2022fast}, large-scale environments~\cite{NeRFLarge:turki2022mega,NeRFLarge:xiangli2022bungee,NeRFLarge:tancik2022block}, and scene understanding~\cite{NeRFLayer:zhi2021inplace,NeRFLayer:ye2022Intrinsic}. Specifically, Li~\etal~\cite{NeRFDynamic:li2021neural} model scene motion through forward and backward flow fields, while Fang~\etal~\cite{NeRFDynamic:fang2022fast} use time-aware voxel features. For large environments, Mega-NeRF and Block-NeRF decompose scenes into local radiance fields, whereas BungeeNeRF progressively grows its representation. Semantic extensions jointly encode scene labels and appearance, and intrinsic models separate reflectance, shading, and residual components. Neural fields also support 3D-aware content generation~\cite{NeRFGAN:gu2021style,NeRFGAN:deng2022gram,NeRFGAN:chen2023sd}. However, high-quality rendered images do not ensure accurate geometry: the surfaces extracted from the underlying density field can still contain noise and inaccuracies~\cite{NeRFShape:yariv2021volume,NeRFBake:yariv2023bakedsdf}. We therefore focus on improving implicit geometry through neural rendering.

\subsection{Reconstruction through Neural Rendering}

Recovering geometry from images remains under-constrained because different combinations of shape and radiance can explain the same observations~\cite{NeRF:zhang2020nerf++,NeRF:barron2022mip}. Existing methods address this ambiguity through surface representations, sampling strategies, geometric priors, and multi-view constraints.

One approach is to make the surface representation more explicit. Niemeyer~\etal~\cite{NeRFShape:niemeyer2020DVR} proposed Differentiable Volumetric Rendering (DVR), which models shape with an occupancy network and differentiates the surface projection through implicit differentiation. VolSDF~\cite{NeRFShape:yariv2021volume} and NeuS~\cite{NeRFShape:wang2021neus} instead optimize SDF-based surfaces through volume rendering. DVR uses object masks during training, while SDF-based methods such as VolSDF and NeuS can reconstruct surfaces without mask supervision. In NeuS, the volume-rendering formulation reduces the geometric bias in the recovered zero-level set. Appropriate initialization and a bounding-sphere prior also guide its surface optimization~\cite{NeRFShape:wang2021neus}. UNISURF~\cite{NeRFShape:Oechsle2021unisurf} unifies implicit surfaces and radiance fields and progressively narrows the sampling region around the estimated surface, allowing reconstruction without input masks.

Hybrid representations combine neural fields with explicit spatial structures. Neural Sparse Voxel Fields (NSVF)~\cite{NeRFVox:liu2020neural} organizes local implicit fields in sparse voxels, which also restrict the sampling region during ray marching. Its sparse voxel structure avoids evaluating empty regions, and progressive optimization supports the recovery of finer scene geometry. Munkberg~\etal~\cite{NeRFTet:munkberg2022extracting} optimize a discrete SDF on a deformable tetrahedral grid and render the resulting mesh during training. Their framework estimates geometry, spatially varying materials, and lighting from posed images and object masks. For efficient rendering, MobileNeRF~\cite{NeRFBake:chen2022mobile} encodes appearance and opacity on polygon faces and bakes the result into textures, while BakedSDF~\cite{NeRFBake:yariv2023bakedsdf} optimizes a hybrid surface-volume representation for unbounded scenes and bakes it for rasterization. Such approaches modify the representation or rendering pipeline to connect neural reconstruction with explicit geometry.

Besides the representation, geometric regularization and additional cues can improve reconstruction. NeRF~\cite{NeRF:Mildenhall2021} injects noise into density predictions to regularize ambiguous regions. UNISURF~\cite{NeRFShape:Oechsle2021unisurf} encourages local normal smoothness, and PlenOctrees~\cite{NeRFRender:yu2021plenoctrees} uses a sparsity prior to favor empty space in ambiguous regions. MonoSDF~\cite{NeRFShape:yu2022monosdf} incorporates monocular depth and normal estimates to constrain the surface. Based on predictions from general-purpose monocular estimators, these cues improve reconstruction quality and reduce optimization time. Guo~\etal~\cite{NeRFShape:guo2022manhattan} apply planar constraints to wall and floor regions based on the Manhattan-world assumption. Semantic segmentation identifies these regions, and the constraints encourage their surface normals to follow parallel or orthogonal directions. Other approaches use direct geometric observations: Rematas~\etal~\cite{NeRFShape:rematas2022urban} use lidar supervision to constrain reconstruction and limit ray-marching samples to a bounded region. These constraints reduce ambiguity by imposing smoothness, sparsity, or prior geometric information.

Multi-view methods obtain geometric cues from agreement between captured images. MVSNeRF~\cite{NeRFMVS:chen2021mvsnerf} builds plane-swept cost volumes for geometry-aware scene reasoning and combines them with volume rendering to reconstruct a radiance field from a small set of nearby views. NerfingMVS~\cite{NeRFMVS:wei2021nerfingmvs} uses learned depth priors, adapted with sparse structure-from-motion observations, to guide NeRF optimization. NeuralWarp~\cite{NeRFMVS:darmon2022improving} combines image warping with volume rendering to optimize photo-consistency across source views. These methods use captured images or depth priors to constrain the geometry through multi-view observations. Direct photo-consistency methods must also account for source-view selection and visibility when comparing projections of the same 3D point.

In this paper, we constrain the density field of vanilla NeRF through agreement among virtual rays rendered by the same network. Specifically, the cross-ray constraint compares appearance predictions around each candidate point, with visibility weights reducing the influence of occluded rays. This provides a self-supervised geometric constraint without image warping or additional depth, normal, or correspondence supervision in the training losses.

\section{Cross-Ray Consistency for Implicit Geometry Optimization}

\begin{figure}[htbp]
    \begin{center}
        \includegraphics[width=\linewidth]{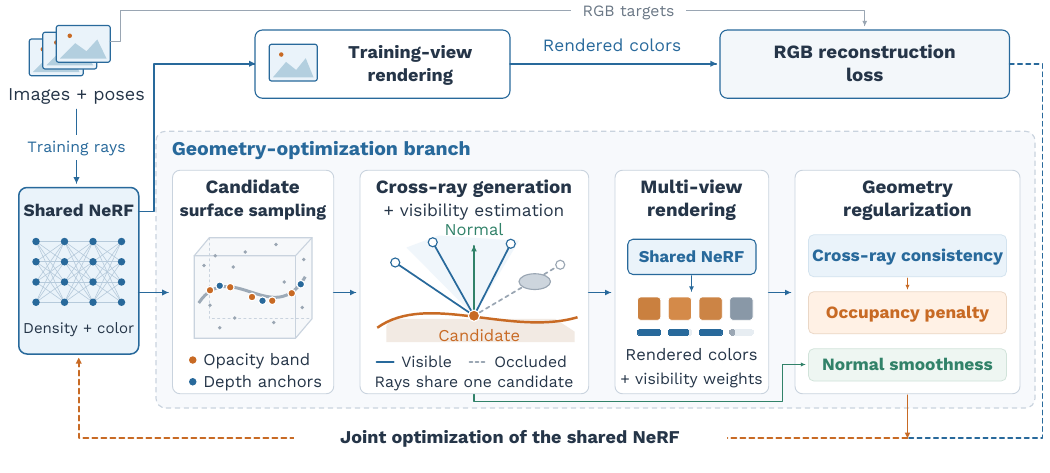}
    \end{center}
    \caption{Overview of the proposed framework. The shared NeRF renders training views for RGB reconstruction and supplies the density field for candidate surface sampling. The geometry-optimization branch generates virtual cross-rays through each candidate, renders their appearances, and estimates visibility weights to reduce the influence of occluded rays. Cross-ray consistency, occupancy suppression, and local normal smoothness jointly regularize the field.}
    \Description{A shared NeRF with an RGB reconstruction path and a geometry-optimization branch. The branch samples candidate points, generates rays that intersect at a common candidate, renders their colors with the shared network, and applies visibility-aware consistency, occupancy suppression, and normal smoothness. Both paths update the same NeRF.}
    \label{fig:framework}
\end{figure}

Our framework extends vanilla NeRF with a geometry-optimization branch, as shown in Figure~\ref{fig:framework}. The NeRF network still predicts density and view-dependent color from 3D positions and viewing directions. The added branch samples candidate surface points from the current density field, generates virtual viewpoints around them, and performs multi-view appearance rendering for rays passing through each point. Based on these predictions, we impose visibility-aware cross-ray consistency and geometric regularization to optimize the field.

\subsection{Overview and Cross-Ray Consistency}
\label{formulation}

A vanilla NeRF represents a scene with a function $F_\theta$ that maps a position $\mathbf{x}\in\mathbb{R}^3$ and viewing direction $\mathbf{d}\in\mathbb{R}^3$ to density $\sigma_\theta(\mathbf{x})$ and color $\mathbf{c}_\theta(\mathbf{x},\mathbf{d})$:
\begin{equation}
    F_\theta(\mathbf{x},\mathbf{d})=
    \bigl(\sigma_\theta(\mathbf{x}),\mathbf{c}_\theta(\mathbf{x},\mathbf{d})\bigr).
    \label{equ:rfield}
\end{equation}
For a camera ray $\mathbf{r}(t)=\mathbf{o}+t\mathbf{d}$ with ordered samples $\{t_k\}_{k=1}^{K}$, volume rendering gives
\begin{align}
    \widehat{\mathbf{C}}(\mathbf{r}) &= \sum_{k=1}^{K}T_k\alpha_k\mathbf{c}_\theta(\mathbf{r}(t_k),\mathbf{d}),
    \label{equ:render}\\
    \alpha_k &= 1-\exp\bigl(-\sigma_\theta(\mathbf{r}(t_k))\delta_k\bigr),
    \qquad
    T_k = \prod_{l<k}(1-\alpha_l),
    \label{equ:alpha_transmittance}
\end{align}
where $\delta_k=t_{k+1}-t_k$. The NeRF rendering branch minimizes the image reconstruction loss
\begin{equation}
    L_{\mathrm{rgb}}=
    \sum_{\mathbf{r}\in\mathcal{R}}
    \left\|\widehat{\mathbf{C}}(\mathbf{r})-\mathbf{C}(\mathbf{r})\right\|_2^2.
    \label{equ:rgb_loss}
\end{equation}

Multi-view reconstruction commonly compares the projections of the same 3D point in captured images. Such methods must select source views, establish correspondences, and handle visibility~\cite{NeRFMVS:darmon2022improving}. Our added branch instead generates a local bundle of virtual rays for each candidate surface point and renders every ray with the NeRF network. These rays intersect at the same point but differ in origin and direction. If the candidate lies on a stable surface, its appearance should remain similar across the visible rays in this local bundle. We therefore use their color disagreement as a self-supervised constraint, without image warping or ground-truth depth.

\subsection{Candidate Surface Sampling}
\label{candidate_sampling}

\begin{figure}[htbp]
    \centering
    \includegraphics[width=\linewidth]{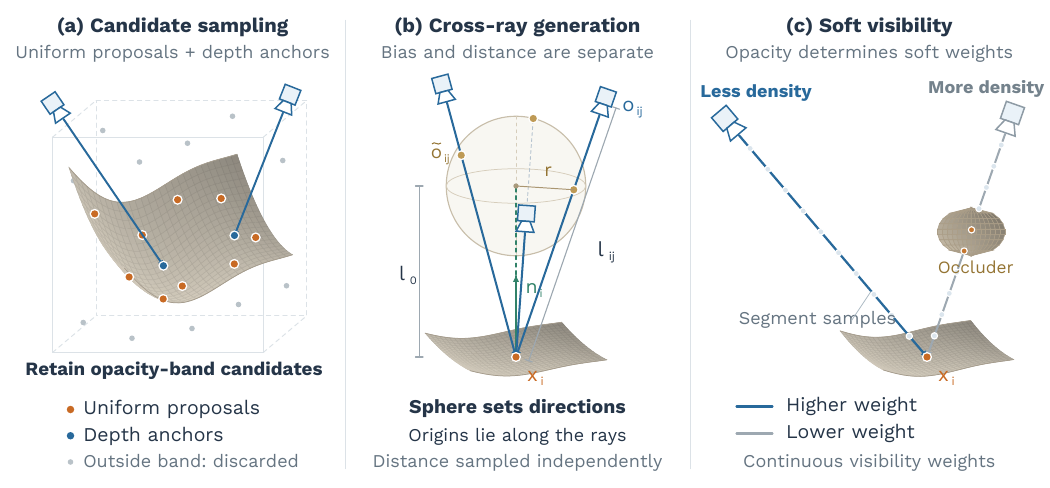}
    \caption{Details of candidate sampling and visibility-aware cross-ray construction. (a) Uniform proposals and rendered-depth anchors are filtered by an intermediate opacity band to obtain candidate points. The curved patch schematically represents the current density boundary. (b) For a candidate $\mathbf{x}_i$, the auxiliary sphere is centered at $\mathbf{x}_i+l_0\mathbf{n}_i$ with radius $r$. Its sampled locations $\widetilde{\mathbf{o}}_{ij}$ determine ray directions, while the actual origins $\mathbf{o}_{ij}$ are placed independently at distances $l_{ij}$. All generated rays pass through the same candidate. (c) Opacity samples along each origin-to-candidate segment determine a continuous visibility weight: an intervening occupied region increases the opacity sum and reduces the ray's influence. This segment is used for visibility estimation; appearance is obtained by standard volume rendering along the generated ray.}
    \Description{Three original schematic panels. Candidate sampling combines orange uniform samples near a curved density boundary with blue rendered-depth anchors; gray proposals outside the opacity band are discarded. Cross-ray generation uses a sphere shifted along a green outward normal to determine directions; blue camera symbols show actual origins at independently sampled distances. Visibility estimation compares two rays through the same candidate, with sample points on their paths. An intervening occupied region produces a smaller soft weight, while the clearer path receives a larger weight.}
    \label{fig:method_details}
\end{figure}

As illustrated in Figure~\ref{fig:method_details}(a), we combine random proposals with rendered-depth anchors to obtain candidate surface points. We first sample random points in a bounded scene volume $\mathcal{V}$ and query the current density field. With a fixed interval $\delta$, we convert density to opacity,
\begin{equation}
    \alpha(\mathbf{x})=1-\exp\bigl(-\sigma_\theta(\mathbf{x})\delta\bigr).
    \label{equ:point_opacity}
\end{equation}
We then select candidate points from an intermediate opacity band:
\begin{equation}
    \mathcal{S}_{\mathrm{rand}}=
    \left\{\mathbf{x}\in\mathcal{V}\mid
    \alpha_{\min}<\alpha(\mathbf{x})<\alpha_{\max}\right\},
    \qquad
    \alpha_{\min}=0.5,\quad \alpha_{\max}=0.95.
    \label{equ:random_candidates}
\end{equation}
We retain intermediate-opacity samples as surface candidates, excluding low-density samples and those with nearly saturated opacity.

To complement the random candidates, we also obtain surface anchors from rendered training rays. Specifically, using the rendering weights in Equation~\eqref{equ:render}, we define the unnormalized rendered-depth estimate $\widehat{t}(\mathbf{r})=\sum_k T_k\alpha_k t_k$. The corresponding anchor is
\begin{equation}
    \mathbf{x}_{\mathrm{ray}}=
    \mathbf{o}+\widehat{t}(\mathbf{r})\mathbf{d},
    \qquad
    \mathcal{S}=\mathcal{S}_{\mathrm{rand}}\cup\mathcal{S}_{\mathrm{ray}}.
    \label{equ:ray_anchors}
\end{equation}
We apply the same opacity-band test to the rendered-depth anchors before adding them to $\mathcal{S}_{\mathrm{ray}}$. Based on the combined set $\mathcal{S}$, we randomly retain a candidate batch for the geometry branch. The uniform proposals explore the current field, while the rendered-depth anchors provide candidates at surface locations estimated along the training rays.

\subsection{Cross-Ray Generation and Visibility Estimation}
\label{crossray_generation}

Figure~\ref{fig:method_details}(b) illustrates the cross-ray construction. For each candidate $\mathbf{x}_i\in\mathcal{S}$, we first estimate a local surface direction from the opacity gradient:
\begin{equation}
    \mathbf{n}_i=
    -\frac{\nabla_{\mathbf{x}}\alpha(\mathbf{x}_i)}
    {\left\|\nabla_{\mathbf{x}}\alpha(\mathbf{x}_i)\right\|_2+\epsilon}.
    \label{equ:opacity_normal}
\end{equation}
This normal is computed on the fly from the radiance field rather than supplied as ground truth. We keep its computation in the automatic-differentiation graph so that cross-ray construction and normal smoothness can update the density field. Next, we sample $\mathbf{u}_{ij}\sim\mathcal{U}(\mathbb{S}^2)$ and form an auxiliary location
\begin{equation}
    \widetilde{\mathbf{o}}_{ij}
    =\mathbf{x}_i+l_0\mathbf{n}_i+r\mathbf{u}_{ij}.
    \label{equ:aux_origin}
\end{equation}
Here $l_0$ is a normal-direction bias: it shifts the isotropic perturbation $r\mathbf{u}_{ij}$ toward the estimated outward side of the candidate surface. The auxiliary location only determines the ray direction. Defining the normal-bias strength $\lambda_{\mathrm{n}}=l_0/r$ for $r>0$, we obtain
\begin{equation}
    \mathbf{d}_{ij}=
    \frac{\mathbf{x}_i-\widetilde{\mathbf{o}}_{ij}}
    {\left\|\mathbf{x}_i-\widetilde{\mathbf{o}}_{ij}\right\|_2}
    =-\frac{\lambda_{\mathrm{n}}\mathbf{n}_i+\mathbf{u}_{ij}}
    {\left\|\lambda_{\mathrm{n}}\mathbf{n}_i+\mathbf{u}_{ij}\right\|_2}.
    \label{equ:crossray_direction}
\end{equation}
Thus, the virtual-viewpoint directions depend on $l_0$ and $r$ through their ratio. When $\lambda_{\mathrm{n}}=0$, they are isotropic around the candidate; increasing $\lambda_{\mathrm{n}}$ continuously favors the estimated normal side without explicitly rejecting sampled directions. We independently draw $l_{ij}\sim\mathcal{U}(l_{\min},l_{\max})$ to place the final ray origin:
\begin{equation}
    \mathbf{o}_{ij}=\mathbf{x}_i-l_{ij}\mathbf{d}_{ij},
    \qquad
    \mathbf{r}_{ij}(t)=\mathbf{o}_{ij}+t\mathbf{d}_{ij}.
    \label{equ:crossray}
\end{equation}
The ratio $\lambda_{\mathrm{n}}$ therefore controls the angular normal bias, whereas $l_{ij}$ controls the actual ray-origin distance. By construction, all rays indexed by $j$ intersect at $\mathbf{x}_i$. Standard volume rendering in Equation~\eqref{equ:render} gives their colors $\mathbf{c}_{ij}=\widehat{\mathbf{C}}(\mathbf{r}_{ij})$.

However, a generated ray may pass through another occupied region before reaching the candidate, as shown in Figure~\ref{fig:method_details}(c). To reduce the influence of such rays, we sample $M$ points $\{\mathbf{p}_{ijk}\}_{k=1}^{M}$ on the segment from $\mathbf{o}_{ij}$ to $\mathbf{x}_i$ and compute
\begin{align}
    \alpha_{ijk} &= 1-\exp\bigl(-\sigma_\theta(\mathbf{p}_{ijk})\Delta_{ijk}\bigr),\\
    m_{ij} &= \sum_{k=1}^{M}\alpha_{ijk},\\
    w_{ij} &= \frac{1}{1+\exp\bigl(\beta(m_{ij}-\gamma)\bigr)},
    \qquad \beta=20,\quad \gamma=1.10.
    \label{equ:visibility_weight}
\end{align}
Here $m_{ij}$ is an accumulated opacity proxy, defined as the sum of per-sample opacities rather than physical transmittance. Based on this proxy, the sigmoid assigns a smaller weight to rays whose paths contain substantial density.
We treat the estimated visibility weights as constants during back-propagation (stop gradients through $w_{ij}$), preventing the consistency loss from directly adjusting them to downweight rays with inconsistent color predictions.

\subsection{Geometry Regularization}
\label{geometry_regularization}

To measure the appearance agreement of the visible rays, let $D_i=\max(\sum_j w_{ij},\eta)$ with $\eta=0.1$ for candidate $i$. We then compute the visibility-weighted mean color and cross-ray variance as
\begin{align}
    \boldsymbol{\mu}_i &=
    \frac{\sum_j w_{ij}\mathbf{c}_{ij}}
    {D_i},\\
    V_i &=
    \frac{\sum_j w_{ij}
    \left\|\mathbf{c}_{ij}-\boldsymbol{\mu}_i\right\|_2^2}
    {D_i}.
    \label{equ:crossray_variance}
\end{align}
The cross-ray consistency loss is the average variance over $N$ candidates:
\begin{equation}
    L_{\mathrm{CRC}}=\frac{1}{N}\sum_{i=1}^{N}V_i.
    \label{equ:crc_loss}
\end{equation}

We assume that a valid surface hypothesis should exhibit similar appearance across visible virtual views within the sampled local angular range. A large cross-view rendering discrepancy therefore provides negative evidence for occupancy at the candidate point. Minimizing color variance alone may change the appearance predictions while leaving unreliable density in place. We therefore use the visibility-weighted variance to penalize the candidate point's opacity:
\begin{equation}
    L_{\mathrm{occ}}=
    \frac{1}{N}\sum_{i=1}^{N}
    \operatorname{sg}(V_i)\alpha(\mathbf{x}_i),
    \label{equ:occupancy_loss}
\end{equation}
where $\operatorname{sg}(\cdot)$ denotes stop-gradient. The variance acts as a fixed weight in this term: a larger $V_i$ gives a stronger penalty on $\alpha(\mathbf{x}_i)$. Stop-gradient prevents the penalty from being reduced by changing $V_i$ through this path.

Although $L_{\mathrm{CRC}}$ is expressed through rendered colors, it also affects geometry. Specifically, each rendered color depends on density through opacity and transmittance in Equation~\eqref{equ:render}, and the cross-ray directions depend on differentiable opacity gradients. In addition, Equation~\eqref{equ:occupancy_loss} directly penalizes the opacity of inconsistent candidates. The opacity-band test itself is discrete and provides no gradient; instead, it updates the candidate distribution as the field changes. Meanwhile, $L_{\mathrm{rgb}}$ keeps the predictions anchored to the captured images during refinement. This reasoning assumes that surface appearance varies modestly across the sampled local directions; strong specular effects may therefore penalize valid geometry.

To further reduce high-frequency noise in the density field, we apply local normal smoothness. Given a random perturbation $\boldsymbol{\varepsilon}_i\sim\mathcal{U}(-s/2,s/2)^3$ and $\mathbf{x}'_i=\mathbf{x}_i+\boldsymbol{\varepsilon}_i$,
\begin{equation}
    L_{\mathrm{smooth}}=
    \frac{1}{N}\sum_{i=1}^{N}
    \left\|\mathbf{n}(\mathbf{x}_i)-\mathbf{n}(\mathbf{x}'_i)\right\|_2.
    \label{equ:smooth_loss}
\end{equation}
Here, \(s>0\) denotes the side length of the cubic neighborhood used to sample local perturbations.

\subsection{Training Objective}
\label{training_objective}

We combine RGB reconstruction from the NeRF path with the constraints from the geometry branch in the training objective
\begin{equation}
    L = L_{\mathrm{rgb}}
    +\lambda_{\mathrm{cr}}
    \left(L_{\mathrm{CRC}}+\lambda_{\mathrm{occ}}L_{\mathrm{occ}}\right)
    +\lambda_{\mathrm{sm}}L_{\mathrm{smooth}}.
    \label{equ:total_loss}
\end{equation}
Here $L_{\mathrm{rgb}}$ maintains agreement with the captured images, while $L_{\mathrm{CRC}}$ supplies the cross-ray constraint. The occupancy and smoothness terms act on the geometry in complementary ways: one suppresses inconsistent density, and the other reduces local noise.

Algorithm~\ref{alg:geometry_update} summarizes one optimization update with a nonempty candidate batch. Specifically, the geometry branch uses the current field to select candidates, construct cross-rays, and render their colors. Based on these predictions, it evaluates the visibility-weighted losses and updates the shared network together with RGB reconstruction. The normal computation and appearance rendering remain differentiable, whereas the visibility weights and the variance used in the occupancy penalty are detached as described above.

\begin{algorithm}[htbp]
    \caption{One update with visibility-aware cross-ray consistency}
    \label{alg:geometry_update}
    \begin{algorithmic}[1]
        \REQUIRE Shared NeRF $F_\theta$, training-ray batch $\mathcal{R}$, sampling volume $\mathcal{V}$, candidate budget $N_{\max}$, cross-ray count $J$
        \STATE Render $\mathcal{R}$ to obtain $L_{\mathrm{rgb}}$ and rendered-depth anchors (Equations~\eqref{equ:rgb_loss} and~\eqref{equ:ray_anchors}).
        \STATE Draw uniform proposals from $\mathcal{V}$ and append the rendered-depth anchors.
        \STATE Apply the opacity-band test and randomly retain up to $N_{\max}$ candidates $\{\mathbf{x}_i\}_{i=1}^{N}$.
        \FOR{each candidate $i=1,\ldots,N$}
            \STATE Compute the differentiable opacity normal $\mathbf{n}_i$ (Equation~\eqref{equ:opacity_normal}).
            \FOR{each virtual ray $j=1,\ldots,J$}
                \STATE Draw $\mathbf{u}_{ij}\sim\mathcal{U}(\mathbb{S}^2)$ and $l_{ij}\sim\mathcal{U}(l_{\min},l_{\max})$.
                \STATE Construct the normal-biased direction $\mathbf{d}_{ij}$ (Equation~\eqref{equ:crossray_direction}).
                \STATE Set $\mathbf{o}_{ij}=\mathbf{x}_i-l_{ij}\mathbf{d}_{ij}$ and $\mathbf{r}_{ij}(t)=\mathbf{o}_{ij}+t\mathbf{d}_{ij}$.
                \STATE Render $\mathbf{c}_{ij}=\widehat{\mathbf{C}}(\mathbf{r}_{ij})$ by Equation~\eqref{equ:render}.
                \STATE Sum sampled opacities on the origin-to-candidate segment to obtain $m_{ij}$.
                \STATE Set $w_{ij}=\operatorname{sg}\!\left([1+\exp(\beta(m_{ij}-\gamma))]^{-1}\right)$.
            \ENDFOR
            \STATE Compute $\boldsymbol{\mu}_i$ and $V_i$ using $w_{ij}$ (Equation~\eqref{equ:crossray_variance}).
        \ENDFOR
        \STATE Compute $L_{\mathrm{CRC}}$, $L_{\mathrm{occ}}=N^{-1}\sum_i\operatorname{sg}(V_i)\alpha(\mathbf{x}_i)$, and $L_{\mathrm{smooth}}$.
        \STATE Form the training objective in Equation~\eqref{equ:total_loss} and update $\theta$.
        \ENSURE Updated shared NeRF $F_\theta$
    \end{algorithmic}
\end{algorithm}

\section{Experiments} \label{sec:experiments}

\subsection{Implementation Details}

For the reported experiments, we initialize the NeRF network from a vanilla-NeRF checkpoint and continue optimization with the objective in Equation~\eqref{equ:total_loss}. In each geometry-optimization step, we draw 20,000 uniform proposals from the sampling volume, append rendered-depth anchors, apply the opacity-band filter, and randomly retain up to 100 candidates. We generate 15 cross-rays for each candidate. For the opacity and normal computations, we use $\delta=(10.0-0.1)/128$ and $\epsilon=10^{-5}$, respectively. Unless noted below, we set $r=0.15$ and $l_0=0.28$, giving a default normal-bias strength of $\lambda_{\mathrm{n}}=l_0/r\approx1.87$. We also set $(l_{\min},l_{\max})=(0.15,0.45)$, $s=0.08$, $\lambda_{\mathrm{cr}}=1.5\times10^{-3}$, $\lambda_{\mathrm{occ}}=1.15$, and $\lambda_{\mathrm{sm}}=2.5\times10^{-4}$.

For visibility estimation, we place 32 uniform samples between $0.01l_{ij}$ and $l_{ij}$ along each origin-to-candidate segment, including the endpoint, and set $\Delta_{ijk}=0.99l_{ij}/32$. Cross-ray rendering uses near and far bounds of 0.1 and 10.0 meters. The continuation with the geometry branch starts at a learning rate of $1.15\times10^{-4}$.

Room~1 uses $(l_{\min},l_{\max})=(0.125,0.30)$, $\lambda_{\mathrm{occ}}=0.8$, and $\lambda_{\mathrm{sm}}=2.5\times10^{-3}$. Office~2 uses $r=0.25$, $(l_{\min},l_{\max})=(0.08,0.70)$, $s=0.10$, $\lambda_{\mathrm{cr}}=1.05\times10^{-3}$, and $\lambda_{\mathrm{occ}}=3.25$; its scene-specific continuation starts at a learning rate of $1.25\times10^{-4}$. These settings are fixed in the reported runs.

For Office~2, we additionally suppress residual floating density with an auxiliary global sparsity penalty:
\begin{equation}
    L_{\mathrm{sparse}}=
    1-\mathbb{E}_{\mathbf{x}\sim\mathcal{V}}
    \left[\exp\left(-\delta_s\operatorname{ReLU}
    (\sigma_\theta(\mathbf{x}))\right)\right].
    \label{equ:sparse_loss}
\end{equation}
The optimization objective for this scene is $L+\lambda_{\mathrm{sp}}L_{\mathrm{sparse}}$, where $L$ is defined in Equation~\eqref{equ:total_loss}, $\lambda_{\mathrm{sp}}=10^{-2}$, and $\delta_s=0.01$. This auxiliary term is not used for the other seven scenes.

For the comparison in Table~\ref{tab:chamfer}, both vanilla NeRF and our method are evaluated after 300,000 cumulative training iterations. When our run is initialized from a vanilla-NeRF checkpoint, its continuation iterations are counted within this total. Thus, both methods have the same iteration budget.

All surface normals used during refinement are computed on the fly from opacity gradients; ground-truth normals are not used. For Replica scenes, the reference mesh determines an oriented 3D sampling volume, but does not enter the training losses. Ground-truth depth is reserved for evaluation. Thus, the losses use no ground-truth depth or normal supervision, although the sampling bounds still depend on reference geometry in the current experiments.

\subsection{Geometry Reconstruction Results}

Table~\ref{tab:chamfer} reports the Chamfer distance on eight Replica scenes at the matched 300,000-iteration budget. Compared with vanilla NeRF, our method improves six scenes and reduces the mean distance by 30.8\%. Room~1 and Office~0 are the two scenes for which the distance increases.

\begin{table}[htbp]
    \centering
    \caption{Chamfer distance on Replica scenes after 300,000 cumulative training iterations for both methods, multiplied by 100 and rounded to two decimal places (lower is better). Relative changes are computed from the unrounded distances with respect to vanilla NeRF; downward arrows indicate reductions. The lower distance in each scene and the entire mean row are highlighted in bold.}
    \label{tab:chamfer}
    \begin{tabular}{lccc}
        \toprule
        Scene & Baseline (Vanilla NeRF) & Ours & Relative change \\
        \midrule
        Room 0   & 0.49 & \textbf{0.29} & $\boldsymbol{\downarrow}$ \textbf{40.0\%} \\
        Room 1   & \textbf{0.54} & 0.65 & $\uparrow$ 21.2\% \\
        Room 2   & 0.67 & \textbf{0.22} & $\boldsymbol{\downarrow}$ \textbf{67.4\%} \\
        Office 0 & \textbf{0.41} & 0.48 & $\uparrow$ 16.2\% \\
        Office 1 & 2.78 & \textbf{2.43} & $\boldsymbol{\downarrow}$ \textbf{12.5\%} \\
        Office 2 & 1.33 & \textbf{0.48} & $\boldsymbol{\downarrow}$ \textbf{64.2\%} \\
        Office 3 & 1.42 & \textbf{0.85} & $\boldsymbol{\downarrow}$ \textbf{40.4\%} \\
        Office 4 & 0.58 & \textbf{0.29} & $\boldsymbol{\downarrow}$ \textbf{50.5\%} \\
        \midrule
        \textbf{Mean} & \textbf{1.03} & \textbf{0.71} & $\boldsymbol{\downarrow}$ \textbf{30.8\%} \\
        \bottomrule
    \end{tabular}
\end{table}

\begin{figure}[htbp]
    \begin{center}
        \includegraphics[width=0.9\linewidth]{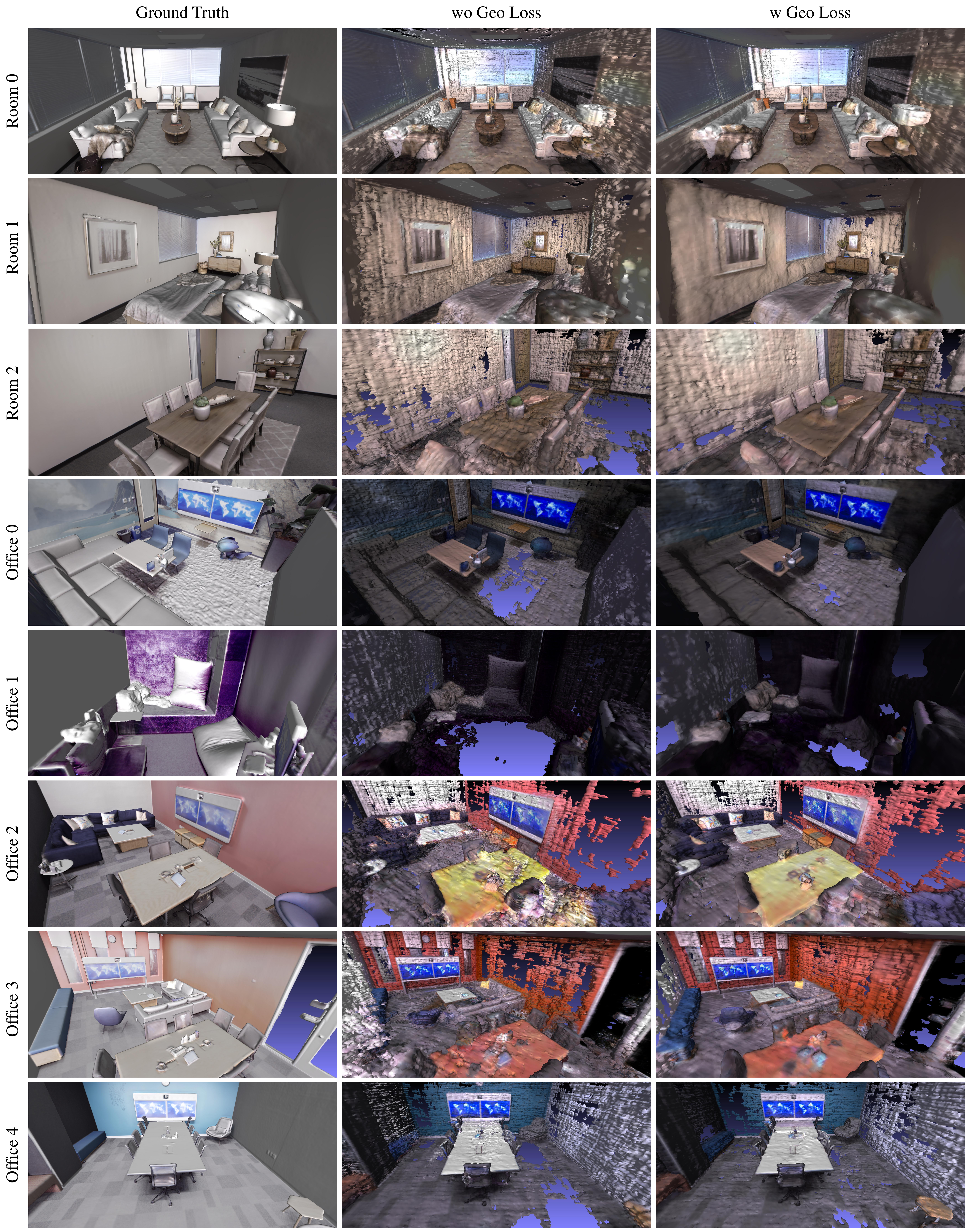}
    \end{center}
    \caption{Qualitative geometry comparison on Replica scenes. From left to right, each row shows the ground-truth geometry, vanilla NeRF, and the reconstruction after optimization with the proposed geometry-refinement objective.}
    \Description{Geometry reconstructions for five office scenes and three room scenes, comparing ground truth, vanilla NeRF, and refinement with the proposed objective.}
    \label{fig:res}
\end{figure}

Figure~\ref{fig:res} qualitatively compares the geometry extracted from the density fields with the ground-truth scene geometry. Although vanilla NeRF recovers the overall scene layout, its surfaces often contain pronounced noise and irregular deformations. After geometry optimization, several wall and furniture surfaces become smoother and more closely resemble the reference geometry. Specifically, the wall in Room~1 and the tabletops in Room~2 and Office~2 exhibit reduced surface irregularities. Together with the quantitative results, these comparisons demonstrate that the proposed geometry optimization can improve the geometric fidelity of NeRF's underlying density field. The smoother wall in Room~1 is a local improvement, despite the higher scene-level Chamfer distance reported in Table~\ref{tab:chamfer}.

\section{Conclusion and Discussion}

We proposed a framework that augments vanilla NeRF with a geometry-optimization branch. Specifically, the branch samples candidate surface points, generates virtual cross-rays through them, and renders their appearance from multiple viewpoints. Based on the visibility-weighted disagreement of these rendered colors, cross-ray consistency constrains the density field, while occupancy suppression and local normal smoothness reduce geometric noise. On eight Replica scenes, the method reduces the mean Chamfer distance by 30.8\% relative to vanilla NeRF.

The current experiments use a reference mesh to define sampling bounds and scene-specific settings for Room~1 and Office~2. Future work should estimate the bounds from training cameras, isolate the effect of the added branch through an RGB-only continuation from the same checkpoint, and investigate appearance models that are less sensitive to specular surfaces.

\bibliographystyle{ACM-Reference-Format}
\bibliography{sample-base}
\end{document}